\documentclass[wcp]{jmlr}

\usepackage{longtable}

\usepackage{algorithm}
\usepackage{algpseudocode}
\usepackage{amsmath}

\usepackage[utf8]{inputenc} 
\usepackage[T1]{fontenc}    
\usepackage{url}            
\usepackage{booktabs}       
\usepackage{amsfonts}       
\usepackage{nicefrac}       
\usepackage{microtype}      
\usepackage{xcolor}         

\usepackage{float}

\usepackage{array}

\usepackage[capitalize]{cleveref}
\crefname{section}{Sec.}{Secs.}
\Crefname{section}{Section}{Sections}
\Crefname{table}{Table}{Tables}
\crefname{table}{Tab.}{Tabs.}

\usepackage{booktabs}

\makeatletter
\let\Ginclude@graphics\@org@Ginclude@graphics 
\makeatother

\jmlrvolume{189}
\jmlryear{2022}
\jmlrworkshop{ACML 2022}

\title[AdvLS]{Adversarial Laser Spot: Robust and Covert Physical-World Attack to DNNs}

  \author{\Name{Chengyin Hu} \Email{1577939987@qq.com}\\
  \addr University of Electronic Science and Technology of China
  \AND
  \Name{Yilong Wang} \Email{w1918566534@163.com}\\
  \addr Chongqing Jiaotong University
  \AND
  \Name{Kalibinuer Tiliwalidi} \Email{975515345@qq.com}\\
  \addr University of Electronic Science and Technology of China
  \AND
  \Name{Wen Li} \thanks{Corresponding author} \Email{liwen@uestc.edu.cn}\\
  \addr University of Electronic Science and Technology of China
 }

\editors{Emtiyaz Khan and Mehmet G\"{o}nen}

\begin{document}

\maketitle

\begin{abstract}

Most existing deep neural networks (DNNs) are easily disturbed by slight noise. However, there are few researches on physical attacks by deploying lighting equipment. The light-based physical attacks has excellent covertness, which brings great security risks to many vision-based applications (such as self-driving). Therefore, we propose a light-based physical attack, called adversarial laser spot (\textbf{\small AdvLS}), which optimizes the physical parameters of laser spots through genetic algorithm to perform physical attacks. It realizes robust and covert physical attack by using low-cost laser equipment. As far as we know, AdvLS is the first light-based physical attack that perform physical attacks in the daytime. A large number of experiments in the digital and physical environments show that AdvLS has excellent robustness and covertness. In addition, through in-depth analysis of the experimental data, we find that the adversarial perturbations generated by AdvLS have superior adversarial attack migration. The experimental results show that AdvLS impose serious interference to advanced DNNs, we call for the attention of the proposed AdvLS. The code of AdvLS is available at: \href{https://github.com/ChengYinHu/AdvLS}{https://github.com/ChengYinHu/AdvLS}.

\end{abstract}

\section{Introduction}
\label{sec1}
The applications based on computer vision are gradually popularized in daily life, such as autonomous vehicle, face recognition system and so on. At the same time, adversarial attack technology has become the focus of many scholars. In the digital environment, adversarial attacks are performed by manipulating pixel-level perturbations \cite{ref3, ref4}, perturbations generated in this setting are invisible to the naked eye. In the physical environment, stickers are attached to the target object as the perturbations to perform adversarial attacks \cite{ref16, ref17, ref18}, which are visible to the naked eye. For example, attaching small pieces of paper to road signs can cause deep neural networks to misclassification, with disastrous results.

In the physical world, there are many natural factors that play the role of imperceptible adversarial perturbations. Such as light, shadow, background environments, etc. If an attacker deliberately mimics the physical adversarial perturbations similar to the natural factor, this physical attack method will inadvertently execute adversarial attacks, resulting in unimaginable consequences. For example, \cite{ref1} introduced a shadow-based physical attacks, which not only ensures the success of physical attacks, but also makes people ignore the existence of the perturbations. Most physical attacks use stickers as physical adversarial perturbations \cite{ref18,ref20}, road signs with stickers, for example, trick deep neural networks. However, these methods have a disadvantage, that is, physical perturbations are always retained on the target objects, so this kind of method has a poor covertness. Some researchers have proposed light-based physical attacks \cite{ref27,ref28}, which make use of the nature of instantaneous attack to ensure the covertness and achieve effective attack. However, these methods usually perform adversarial attacks in dim nighttime environments. In well-lit daytime, they will be completely disabled.

\begin{figure}
    \setlength{\belowcaptionskip}{-0.5cm}  
    \centering
    \includegraphics[width = 0.5\linewidth]{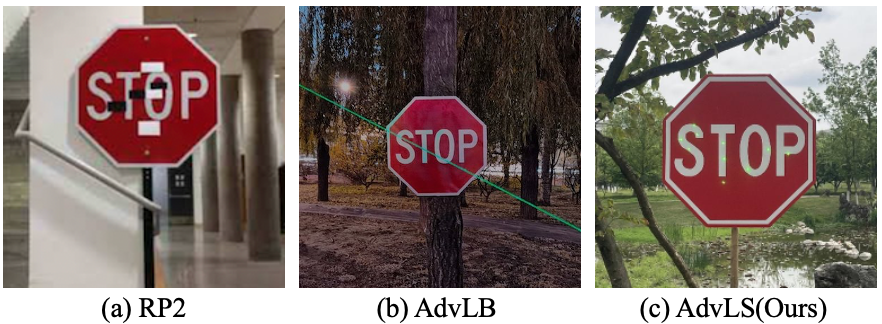}
    \caption{\textbf{Visual comparison}.}
    \label{Figure 1}
\end{figure}

In this paper, we demonstrate a novel light-based physical attack, AdvLS, which uses laser spots as physical perturbations to perform instantaneous attacks on target objects. The advantages of using laser spots to perform physical attack include: (1) Laser spot projection area is puny, with better covertness; (2) Laser performs instantaneous attacks, adversarial perturbations will not be permanently retained on the target objects; (3) Laser spot adversarial attack is currently the first light-based physical attack that perform adversarial attacks in the daytime, making AdvLS more aggressive. Figure \ref{Figure 1} shows a comparison of our method with RP2 \cite{ref18} and AdvLB \cite{ref27}, showing that our approach is much better at covertness.

Our method is simple to execute physical attacks. Firstly, we formalize the physical parameters of laser spots, use genetic algorithm \cite{ref29} to find the physical parameters of the most aggressive laser spots. Finally, based on the physical parameters of laser spots, we use laser pointers to project laser spots to the target objects and generate physical samples. We verify the robustness and covertness of AdvLS through comprehensive experiments in both digital and physical environments, at the same time, some ablation experiments are also presented. Furthermore, by analyzing the misclassification of adversarial samples, we find that the laser spots have some semantic features of clean samples, such as Envelope and Petri dish.

The difficulties of physical attacks mainly include: Print perturbation loss, physical perturbations covertness, robustness, etc. AdvLS uses the nature of light-speed attack to overcome the difficulty of covertness, but also avoids print perturbations. Our main contributions are as follows:

\begin{itemize}
\item We propose a novel physical attack, AdvLS, which is the first light-based method capable of executing physical attacks in the daytime. Our attack equipment is cheap, requires only a set of laser pointers to perform effective physical attacks (See Section \ref{sec1}).

\item We introduce and analyze the existing methods (See Section \ref{sec2}). Then, design strict experimental method and conduct comprehensive experiments to verify the effectiveness of AdvLS (See Section \ref{sec3}, \ref{sec4}). In particular, the light-speed attack nature of laser allows AdvLS to achieve covertness.

\item We conduct a comprehensive analysis of AdvLS, including prediction errors cased by AdvLS, attack migration of AdvLS, etc. These studies will help scholars explore light-based physical attacks (See Section \ref{sec5}). At the same time, we look into some promising mentality for light-based physical attacks (See Section \ref{sec6}).
\end{itemize}

\section{Related work}
\label{sec2}

\subsection{Digital attacks}

\cite{ref2} firstly proposed adversarial attack. After this work, many adversarial attacks were proposed successively \cite{ref13, ref14, ref15}. \cite{ref3} proposed an efficient and simple method called fast gradient sign method (FGSM), which utilizes the gradient information of the model to perform efficient adversarial attacks. \cite{ref4} proposed Deepfool, their work effectively calculated the perturbations that fooled advanced DNNs, thereby reliably quantizing the robustness of many advanced classifiers. \cite{ref5} designed a new loss function, which verified that their adversarial perturbations were more difficult to detect, and pointed out that the inherent characteristics considered as adversarial samples were not in fact. \cite{ref6} proposed an adversarial attack called EAD, experiments showed that EAD could generate adversarial samples with slight distortion and obtain attack effects similar to those of the most advanced methods in different scenes. \cite{ref7} proposed C\&W attack, and proved that defensive distillation network did not significantly improve the robustness of deep neural networks through three attack algorithms. \cite{ref8} proposed an iterative algorithm based on momentum to enhance the adversarial attack, which stabilize the update direction and avoid the local maximum value in the iteration process, generating more transferable adversarial samples. \cite{ref9} improved the migration of adversarial examples by creating different input modes and applying random transformation to input images in each iteration, experiments have shown that the proposed method is more aggressive. \cite{ref10} introduced a new adversarial sample: "Semantic adversarial sample", which first converts RGB images into HSV color space and then randomly moves hue and saturation components while keeping value components constant to generate adversarial samples. \cite{ref11} proposed a black-box adversarial attack based on content, which uses image semantics to selectively modify colors within the selected range of human natural perception, generating unlimited perturbations. \cite{ref12} used human color perception to minimize the disturbance size of perceived color distance and generate adversarial samples.

Different from the above methods, which are executed in the digital environment, physical attacks can not directly modify the input images.

\subsection{Physical attacks}

\cite{ref16} discovered the adversarial samples in the physical world, input the adversarial images obtained by mobile phone camera into the target model. Experimental results showed that, even perceived by the camera, a large proportion of adversarial samples were misclassified. \cite{ref17} disguise adversarial examples in the physical world as reasonable natural styles, which could both fool classifiers and achieve covertness. \cite{ref18} proposed a general attack in the physical world, called RP2, which achieves a robust attack success rate for road sign classifiers in the physical world. \cite{ref19} proposed an adversarial T-shirt, which could avoid the pedestrian detector even if the T-shirt would be deformed with pedestrians. \cite{ref20} proposed a method to create generic, robust and targeted adversarial patches, even if the patches were puny, they would make the target model to ignore other items in the scene. \cite{ref21} designed an adversarial eyeglass frame for attacking face recognition system, and successfully implemented white-box and black-box attacks under different conditions. \cite{ref22} designed adversarial sample with robustness to synthetic noise, distortion and affine transformation, printed the first 3D adversarial sample, proving the existence of robust 3D adversarial samples. \cite{ref23} proposed ShapeShifter, which generate Stop signs of reverse interference, and these signals are always mistaken by classifier as other objects.

Different from the above physical adversarial attacks, light-based physical attacks have better covertness. \cite{ref24} proposed VLA, which is based on visible light, projecting a carefully designed adversarial beam onto the human face to attack the face recognition system. \cite{ref25} studied the real-time physical adversarial attack effect of adversarial light projection on face recognition system, proved the vulnerability of face recognition model to light projection attack. \cite{ref26} used infrared ray as adversarial perturbations to generate adversarial samples, and interpreted the threat of infrared adversarial sample to face recognition system. \cite{ref27} proposed adversarial laser beam (AdvLB), which implements efficient physical adversarial attacks by manipulating the physical parameters of the laser beam. \cite{ref28} introduced an attack system consisting of low-cost projectors, cameras and computers, the proposed attack method can implement effective optical adversarial attack to real 3D objects.

\section{Approach}
\label{sec3}

\subsection{Adversarial sample}

Generating adversarial samples can be regarded as an optimization problem. The input image can be regarded as a high-dimensional vector, in which each element represents a pixel value of the image. Supposing $X$ represents a clean sample, ground truth label $Y$, $f$ represents the classifier, $f(X)$ represents the classification result of picture $X$ by classifier $f$, the classifier $f$ associates with a confidence score ${f}_{Y}(X)$ to class $Y$, ${X}_{adv}$ represents the adversarial sample. the optimization problem can be expressed as:

\begin{equation}
\label{Formula 1}
    f({X}_{adv}) \neq f(X) = Y \quad s.t. \quad ||{X}_{adv} - X|| < \epsilon
\end{equation}

Where, $||\cdot||$represents ${l}_{p}$ norm,  $\epsilon$ represents the threshold of perturbation size. Firstly, the adversarial samples fool the classifier. Secondly, the size of the adversarial perturbation is limited to a certain threshold. 

\begin{figure}
    \setlength{\belowcaptionskip}{-0.5cm}  
    \centering
    \includegraphics[width = 0.6\linewidth]{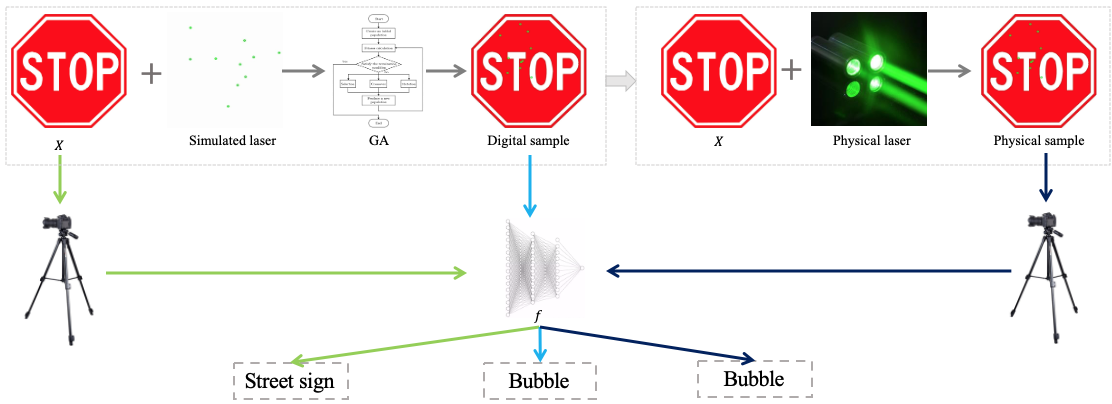}
    \caption{\textbf{Schematic diagram of AdvLS}.}
    \label{Figure 2}
\end{figure}

Figure \ref{Figure 2} shows our method. Firstly, generating simulation laser spots, the genetic algorithm is used to optimize the physical parameters of laser spots and generate adversarial samples in the digital environment. Secondly, manipulating the laser pointers to generate adversarial samples in the physical environment by referring to digital adversarial samples.

\subsection{Genetic algorithm(GA)}

GA \cite{ref29} is a natural heuristic algorithm proposed by John Holland. As the name implies, GA is an algorithm inspired by the genetic and evolution of nature, simulated and implemented on the computer to solve the optimization problems in real life. It is an algorithm that can avoid local optimal solutions.



In this work, we do not use the gradient information of the model, but only the confidence score and prediction label of the model. The advantages of using GA to optimize adversarial samples include:

(1) Simplicity and efficiency: GA is a random optimization algorithm, which does not require excessive mathematical requirements for optimization problems. This algorithm avoids local optimal solutions and find the optimal solution of AdvLS quickly.

(2) Nonlinear problem solving: GA solves the optimization solution of linear problems as well as nonlinear problems. The optimization objective of AdvLS is a nonlinear optimization problem, which can be solved by GA effectively

(3) Requires little information about the target system: GA does not require the optimization problem to be differentiable like the classical optimization problem (e.g. Gradient descent method), which is important in our work, For example, 1) some networks are not differentiable, and 2) computing gradients requires additional information, which in many cases increases the time cost.

\subsection{Laser spot definition}

In this work, we define a laser spot with two parameters: Center position $L (m, n)$ and color $C (r, g, b)$. Each parameter is defined as follows:

\textbf{Center position $L (m, n)$}: $L (m, n)$ represents the center position of the laser spot. We assume that the laser spot is circular, and use a double tuple $(m, n)$ to represent the center position of the laser spot, where $m$ represents the horizontal coordinate of the center, $n$ represents the vertical coordinate.

\textbf{Color $C (r, g, b)$}: $C (r, g, b)$ represents the color of the laser spot. $r$ represents the red channel of the laser color, $g$ represents the green channel, and $b$ represents the blue channel. In the digital environment, laser spots of any color can be generated to carry out adversarial attacks. In the physical environment, due to the limitation of conditions, we choose the green $C (0,255,0)$ laser to execute physical attacks.

The parameters $L (m, n)$ and $C (r, g, b)$ form a laser spot: $\theta(L, C)$. Therefore, the definition of the laser spot group can be expressed as ${G}_{\theta}=({\theta}_{1},{\theta}_{2},...,{\theta}_{i})$. We define a simple function $S(X, {G}_{\theta})$ that synthesize clean image with laser spot group to generate adversarial samples, which means to adopt a simple linear image fusion method to fuse clean image $X$ and laser spot group ${G}_{\theta}$:

\begin{equation}
\label{Formula 2}
    {X}_{adv}=S(X, {G}_{\theta})
\end{equation}

Where, ${X}_{adv}$ represents adversarial sample under the perturbation of the laser spot group. In the digital environment, Formula \ref{Formula 2} represents the generation of adversarial samples. In the physical environment, in order to ensure that the laser spot group appears on the target objects, we design a $l$ function to limit the position area of the laser spot group. Therefore, the adversarial sample generation formula in the physical environment is shown in Formula \ref{Formula 3}:

\begin{equation}
\label{Formula 3}
    {X}_{adv}=S(X, l({G}_{\theta}))
\end{equation}

By using the $l$ function, we ensure that the laser spot group position is limited to the region of the target object.

\subsection{Adversarial laser spot}

Our method consists of two parts :(1) Generating adversarial samples by randomly generating laser spots in the digital environment; (2) In the physical environment, optimizing the physical parameters of laser spots with GA, so as to generate simulated adversarial samples, then using laser pointers to generate physical adversarial samples. Our task is to find adversarial laser spot group ${G}_{\theta}$ that can fool the classifier by GA, the projection area of laser spot on the target area is puny, which allows AdvLS to achieve better covertness. Our optimization objective function is defined as formula \ref{Formula 4}:

\begin{equation}
    \setlength{\abovedisplayskip}{3pt}
    \setlength{\belowdisplayskip}{3pt}
    \label{Formula 4}
    \mathop{\arg\min}_{{G}_{\theta}} {f}_{Y}(S(X, l({G}_{\theta}))) 
\end{equation}

${f}_{Y}(S(X, l({G}_{\theta})))$ represents the confidence score of the adversarial sample on the correct label. The smaller the confidence is, the more adversarial the adversarial sample is. The physical parameters of adversarial laser spot group are optimized and searched by GA, the physical parameter ${G}_{\theta}$ is output when adversarial sample fool the classifier.

In the digital environment, we verify the effectiveness of AdvLS by randomly generating adversarial laser spot group to generate adversarial samples. Then, based on GA, design the adversarial attack algorithm in the physical environment.

\begin{algorithm}[]  
	\caption{\label{Algorithm 1}Pseudocode of AdvLS}
	\KwIn{Input $X$, Classifier $f$, Label $Y$, Population size $Seed$, Iterations $Step$, Crossover probability $Pc$, Mutation probability $Pm$;}
	\KwOut{A vector of parameters ${G}_{\theta}^{\star}$;}
	
	Initiation $Seed$, $Step$, $Pc$, $Pm$;\\ 
	Encoding laser spot group ${G}_{\theta}(i)$ randomly;\\
	\For{steps in range(0, $Step$)}{
		\For{for seeds in range(0, $Seed$)}{
		${X}_{adv}(seeds)=S(X,l({G}_{\theta}(seeds)))$;\\
		${f}_{Y}({X}_{adv}(seeds)) \leftarrow (f({X}_{adv}(seeds)); Y)$;\\
		\If{$f({X}_{adv}) \neq Y$}{
		    ${G}_{\theta}^{\star}={G}_{\theta}(seeds)$;\\
		    Output ${G}_{\theta}^{\star}$;\\
		    Exit();
		}
		}
		Selection with ${f}_{Y}({X}_{adv}(seeds))$,
		Crossover with $Pc$,
		Mutation with $Pm$;
	}
	
\end{algorithm}

As shown in Algorithm \ref{Algorithm 1}, AdvLS takes a clean sample $X$, classifier $f$, population size $Seed$, iterations $Step$, crossover probability $Pc$ and mutation probability $Pm$ as input decided by the attacker. Details of the algorithm have been explained in Algorithm \ref{Algorithm 1}. In this work, our selection strategy is to replace the top tenth of individuals with the highest fitness value with the lowest top tenth of individuals (note that the smaller the fitness value is, the more adversarial the individual is). The advantage of using this selection strategy is to weed out the least aggressive individuals and reduce the time cost. In addition, we set crossover rate $Pc$ and mutation rate $Pm$ to 0.7 and 0.1, respectively. Experimental results show that our selection strategy, crossover rate and variation rate achieve efficient optimization solution to the target problem. The optimum physical parameter ${G}_{\theta}^{\star}$ of laser beam group is output finally, which is used for further carrying out adversarial attacks in the physical world.

\section{Experiment}
\label{sec4}

\subsection{Experimental setting}

\begin{figure}[H]
    \setlength{\belowcaptionskip}{-0.5cm}  
    \centering
    \includegraphics[width = 0.4\linewidth]{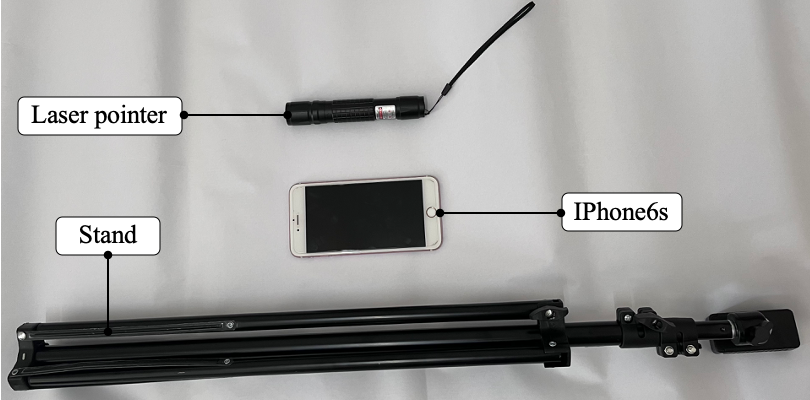}
    \caption{\textbf{Experimental devices}.}
    \label{Figure 4}
\end{figure}

As with the method in AdvLB \cite{ref27}, we use ResNet50 \cite{ref31} as a target model to carry out the adversarial attack experiment in both digital and physical environments. In the digital environment, we randomly selected 1000 correctly classified images from ImageNet \cite{ref32} for testing. In the physical environment, we use street sign, cleaver as target objects for testing. Our experimental devices are shown in Figure \ref{Figure 4}.

In the physical environment, we perform adversarial attacks with laser pointers. We set the number of physical adversarial laser spots to 10, so we need to use 10 laser pointers and 10 tripods, use an iPhone6s as a camera device. It has been verified that different camera devices will not affect the effectiveness of AdvLS. For all tests, we use attack success rate (ASR) as the metric to report the effectiveness of AdvLS.

\begin{figure}[H]
    \centering
    \includegraphics[width = 0.6\linewidth]{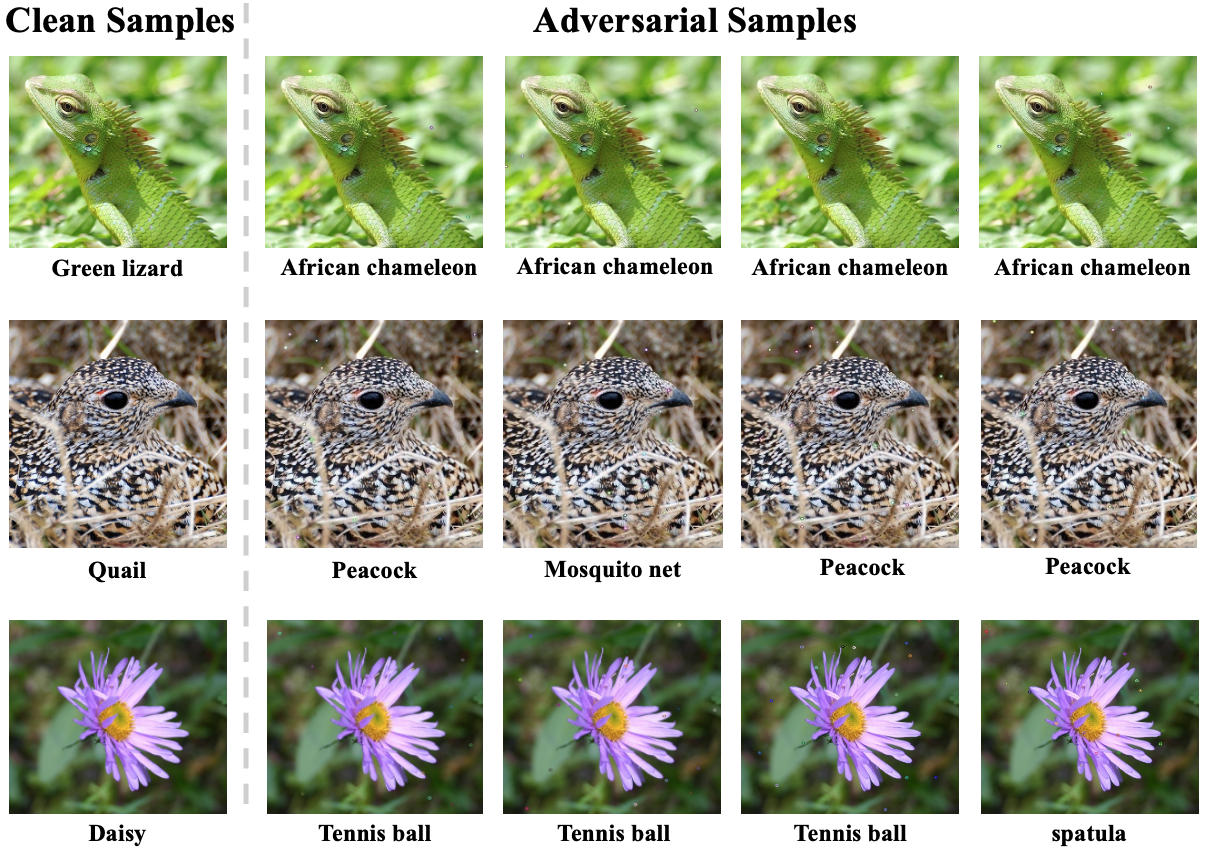}
    \caption{\textbf{Adversarial samples in the digital environment}.}
    \label{Figure 5}
\end{figure}

\subsection{Evaluation of AdvLS}

To ensure the feasibility of AdvLS, we execute experimental tests in the digital environment. Then, the robustness and covertness of AdvLS are verified by physical experiments. 

\textbf{Digital test}: We conduct digital attack experiments on 1000 correctly classified images selected from ImageNet \cite{ref32}. We conduct digital attack experiments on random color, red, green and blue laser spots respectively. The experimental results are shown in Table \ref{Table 1}:

\begin{table}[htbp]
    \setlength{\abovecaptionskip}{0.cm}
    \setlength{\belowcaptionskip}{0.cm}
    \centering
    \caption{\label{Table 1}Attack success rate (ASR) in the digital environment.}
    \begin{tabular}{ccccc}
    \hline
    Color & Random & Red & Green & Blue\\
    \hline
    AdvLS & 75.8\% & 82.6\% & 87.7\% & 78.7\% \\
    \hline
    Query & 237.6 & 204.9 & 143.0 & 241.3 \\
    \hline
    AdvLB & 95.1\% & \O & \O & \O \\
    \hline
    Query & 834.0 & \O & \O & \O \\
    \hline
    \end{tabular}
\end{table}

Table \ref{Table 1} shows the attack success rate and average queries of AdvLS in the digital environment. The number of laser spots ranges from 10 to 50. It can be seen that the adversarial laser spot has a strong antagonism and achieves a robust attack success rate in the digital environment. Even the attack success rate of AdvLS is not as good as AdvLB \cite{ref27}, it's more efficient. In addition, Figure \ref{Figure 5} shows the digital adversarial samples generated by AdvLS, which lead to classifier classification errors by adding a small number of adversarial laser spots that are imperceptible to the naked eye without changing the semantic information of the original image. For example, by adding a few adversarial laser spots to a clean sample, the classifier misclassifies Quail as Peacock, Mosquito net, etc.

\begin{figure}
    \centering
    \includegraphics[width = 1\linewidth]{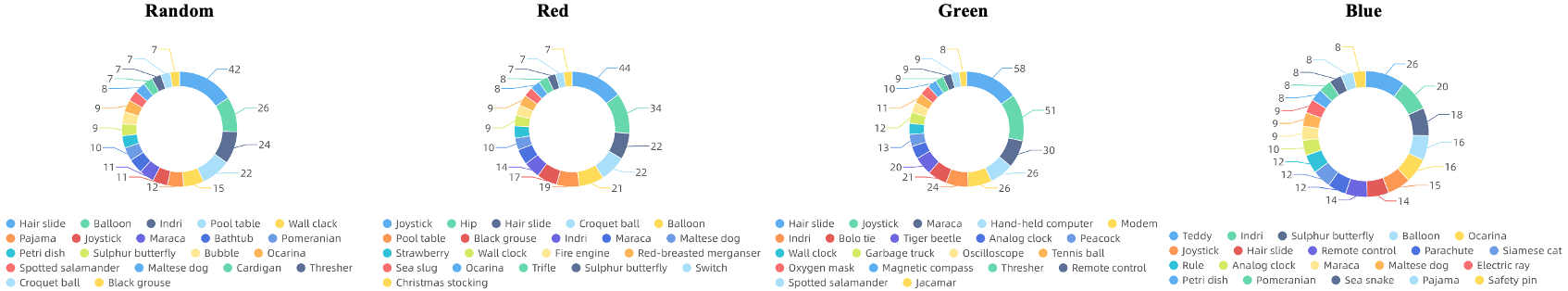}
    \caption{\textbf{Statistics of misclassification in the digital environment}.}
    \label{Figure 6}
\end{figure}

On the other hand, we make statistics on the misclassification results. As shown in Figure \ref{Figure 6}, most of the adversarial samples were misclassified as Hair Slide, Joystick, etc. By looking at the original clean sample of Hair Slide in ImageNet's training set \cite{ref32}, we find that the Hair slide image had many shiny plastic crystals, and these elements were very similar to the effect against laser spots. Some of the same phenomena will be shown in Section \ref{sec5}.

\textbf{Physical test}: In the physical environment, light and shadow affect the accuracy of the classifiers. In order to accurately verify the effectiveness of AdvLS, more carefully designed experiments are conducted. We divide the physical experiments into indoor experiments and outdoor experiments to verify the robustness of AdvLS. In an indoor environment, we perform physical attacks by manipulating the physical parameters of the laser spots. Through a large number of experiments, we achieve a 100\% attack success rate and verified the feasibility of AdvLS in the indoor environment (ASR of 100\% in AdvLB \cite{ref27}). In the outdoor environment, we choose stop sign as the target object. In order to study the attack robustness under different angles, we conduct physical attack from different angles. Through extensive experiments, we verify the robustness of AdvLS. The experimental results are shown in Table \ref{Table 2}.

\begin{table}[htbp]
    \setlength{\abovecaptionskip}{0.cm}
    \setlength{\belowcaptionskip}{-0.cm}
    \centering
    \caption{\label{Table 2}ASR at different angles.}
    \begin{tabular}{cccc}
    \hline
    Angle & ${0}^{\circ}$ & ${30}^{\circ}$ & ${45}^{\circ}$\\
    \hline
    AdvLS & 80.56\% & 77.78\% & 41.67\% \\
    \hline
    AdvLB & 77.40\% & \O & \O \\
    \hline
    \end{tabular}
\end{table}

\begin{figure}
    \setlength{\belowcaptionskip}{-0.5cm}  
    \centering
    \includegraphics[width = 0.4\linewidth]{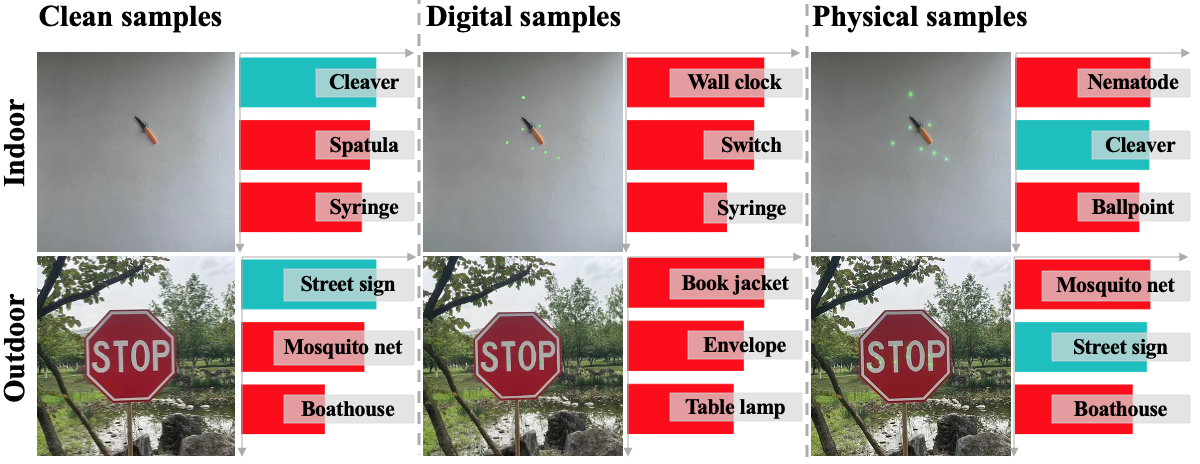}
    \caption{\textbf{Demonstration of adversarial samples from different angles}.}
    \label{Figure 7}
\end{figure}

As can be seen from the experimental results in Table \ref{Table 2}, AdvLS achieves a robuster performance than AdvLB \cite{ref27} in outdoor test. As can be seen from Figure \ref{Figure 7}, the adversarial samples generated by AdvLS have excellent covertness and execute adversarial attacks during the daytime. Note that this is the only light-based physical attack we know of that can be deployed during the daytime. The experimental results in Table \ref{Table 2} and the demonstration of adversarial samples in Figure \ref{Figure 7} verify the robustness and covertness of the proposed AdvLS.

\begin{figure}[H]
    \setlength{\belowcaptionskip}{-0.5cm}  
    \centering
    \includegraphics[width = 0.5\linewidth]{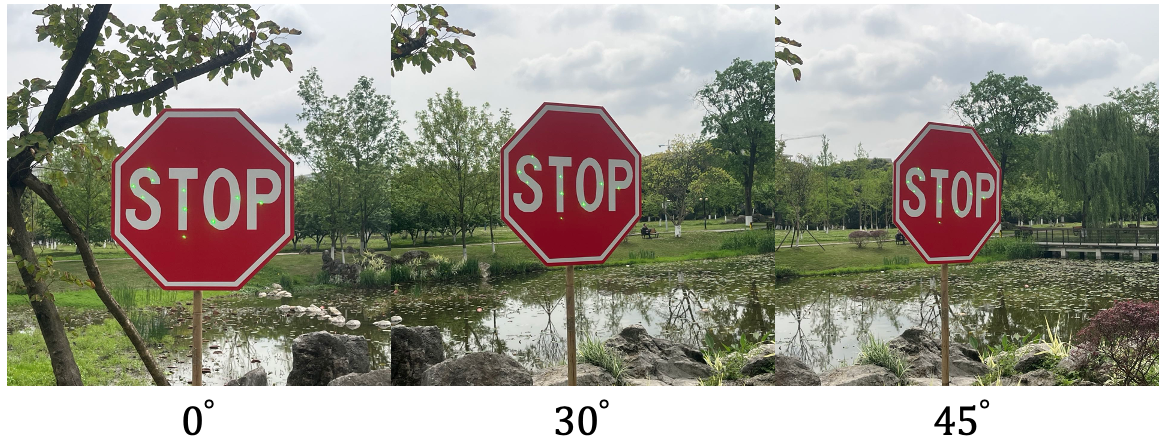}
    \caption{\textbf{ Comparison of digital and physical samples in indoor and outdoor environments}.}
    \label{Figure 8}
\end{figure}

In the physical environment, the simulation laser spots are generated on the computer, the optimal simulation adversarial sample is obtained by GA, then the physical parameters of the adversarial laser spots are saved. Finally, controlling laser pointers to project on the target objects, generate physical adversarial samples. Through the analysis and comparison of simulated samples and physical samples, it can be seen from Figure \ref{Figure 8} that digital samples have an excellent consistency with physical samples.

On the whole, although the ASR of AdvLS is lower than AdvLB in the digital environment, its query efficiency is much better, and AdvLS is significantly more robust than AdvLB in the physical environment. Therefore, comprehensive experiments confirm the effectiveness of our proposed AdvLS in the both digital and physical environments.

\subsection{Ablation study}

In this section, we perform a series of experiments on ImageNet \cite{ref32} to study the adversarial effect of AdvLS with different parameters. The main parameters we study include the number of laser spots and the color of laser spots.

\begin{figure}
    \centering
    \includegraphics[width = 0.4\linewidth]{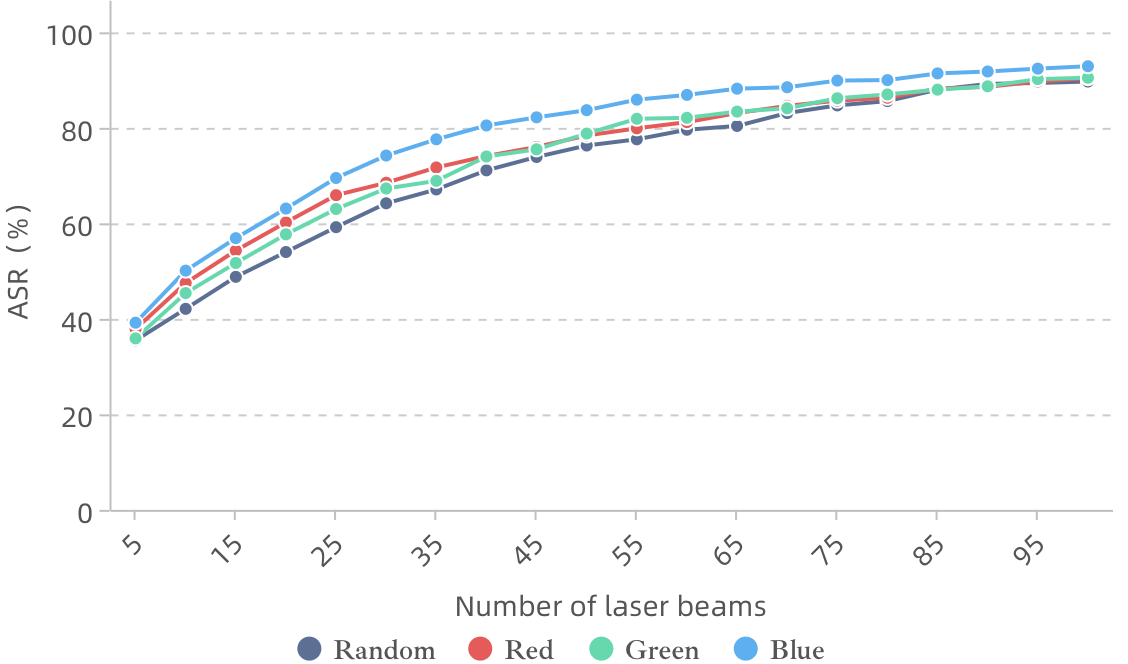}
    \caption{\textbf{ Influence of different color and number of laser spots on AdvLS adversarial effect}.}
    \label{Figure 9}
\end{figure}

In order to study the influence of the number of laser spots on the adversarial effect of AdvLS, we set the value range from 5 to 100 with an interval of 5 for the number of laser spots. As for the color of laser spots, we study the influence of random color, red, green and blue laser spots on AdvLS respectively. The experimental results are shown in Figure \ref{Figure 9}.

The experimental results in Figure \ref{Figure 9} show that: (1) AdvLS achieves a high ASR even with fewer laser spots. When the number of laser spots is 35, it can be seen that AdvLS achieves an ASR about 70\%. Even when we use only 15 laser spots, AdvLS achieves an ASR about 50\%. According to the digital adversarial sample in Figure \ref{Figure 5}, when the number of laser spots is 15, the adversarial perturbations can hardly be detected by naked eyes. (2) Blue laser spots compared with other colors, more antagonistic effect. This phenomenon is consistent with the experimental results in \cite{ref30}.

\section{Discussion}
\label{sec5}

In this section, we discuss some interesting phenomena of AdvLS in both digital and physical environments.

\begin{figure}[H]
    \setlength{\belowcaptionskip}{-0.5cm}  
    \centering
    \includegraphics[width = 0.4\linewidth]{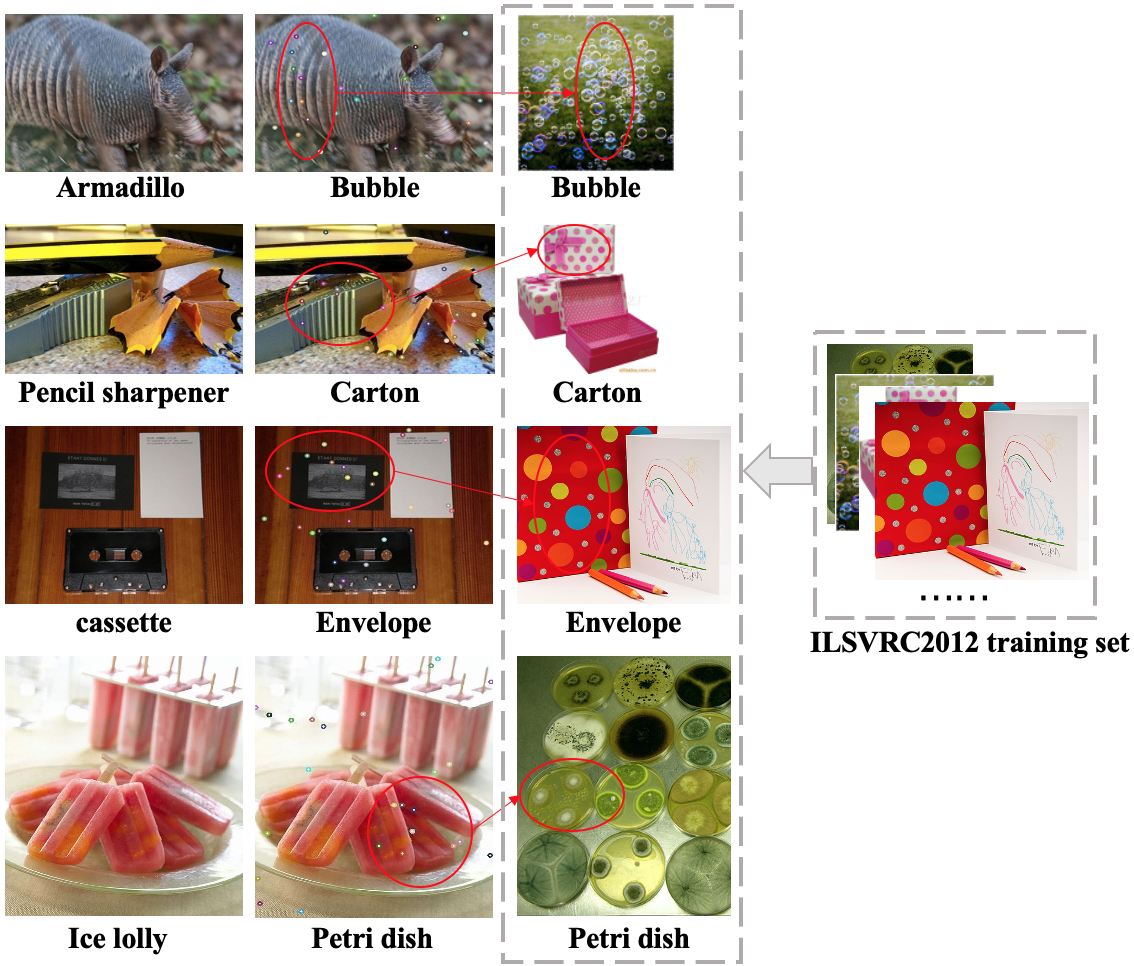}
    \caption{\textbf{ Characteristic analysis of adversarial samples in the digital environment}.}
    \label{Figure10}
\end{figure}

In the digital environment, it can be seen from the experimental results in Table \ref{Table 1} that the adversarial perturbations generated by AdvLS have a robust adversarial effect. As shown in Figure \ref{Figure10}, laser spots contain semantic features of many image categories. By adding adversarial laser spots to clean images, adversarial samples will be misclassified as Bubble, Carton, Envelope, Petri dish, etc. In ImageNet's training set \cite{ref32}, we check the training set samples of Bubble, Carton, Envelope and Petri dish respectively, we find that the adversarial laser spots are very similar to the image features in the training sets. Thus, with the perturbation of a small number of laser spots, the adversarial samples can fool advanced DNNs.

In the physical environment, we manipulate 10 laser Pointers to attack the target objects. In the outdoor environment, a total of 108 physical adversarial samples at various angles are obtained. By analyzing the adversarial samples that could be successfully attacked, we find that the adversarial samples were mainly misclassified as Park bench, Lawn mower, etc. In the indoor environment, a total of 25 adversarial samples are obtained, most of which are misclassified as Modem, Croquet ball, etc.

In addition, we test the adversarial attacks mobility of AdvLS in digital and physical environments. First of all, in the digital environment, the data set is the digital adversarial samples that successfully attack ResNet50 \cite{ref31}, which contains adversarial samples generated by laser spots with random colors, red, green and blue. The experimental results are shown in Table \ref{Table 3}. Secondly, in the physical environment, the data set is the physical adversarial samples that successfully attack ResNet50 \cite{ref31}, which contains the physical adversarial samples of ${0}^{\circ}$, ${30}^{\circ}$ and ${45}^{\circ}$. The experimental results are shown in Table \ref{Table 4}.

\begin{table}[htbp]
    \setlength{\abovecaptionskip}{0.cm}
    \setlength{\belowcaptionskip}{-0.cm}
    \centering
    \caption{\label{Table 3}Attack migration in the digital environment (\%).}
    \begin{tabular}{ccccc}
    \hline
    Classifier & Random & Red & Green & Blue\\
    \hline
    Inception\_V3 \cite{ref38} & 46.0 & 46.9 & 31.5 & 65.4\\
    \hline
    VGG19 \cite{ref34} & 81.3 & 82.0 & 83.2 & 88.3\\
    \hline
    ResNet101 \cite{ref31} & 82.7 & 75.3 & 73.7 & 90.7\\
    \hline
    GoogleNet \cite{ref37} & 63.0 & 62.8 & 55.0 & 81.3\\
    \hline
    AlexNet \cite{ref35} & 95.9 & 97.2 & 96.6 & 97.2\\
    \hline
    DenseNet \cite{ref33} & 64.6 & 67.2 & 54.5 & 75.7\\
    \hline
    MobileNet \cite{ref36} & 92.0 & 91.8 & 87.6 & 96.2\\
    \hline
    \end{tabular}
\end{table}

\begin{table}[htbp]
    \setlength{\abovecaptionskip}{0.cm}
    \setlength{\belowcaptionskip}{-0.cm}
    \centering
    \caption{\label{Table 4}Attack migration in the physical environment (\%).}
    \begin{tabular}{cccc}
    \hline
    Classifier & ${0}^{\circ}$ & ${30}^{\circ}$ & ${45}^{\circ}$\\
    \hline
    Inception\_V3 \cite{ref38} & 10.34 & 64.29 & 80.00\\
    \hline
    VGG19 \cite{ref34} & 20.69 & 39.29 & 66.67\\
    \hline
    ResNet101 \cite{ref31} & 100 & 100 & 86.67\\
    \hline
    GoogleNet \cite{ref37} & 96.55 & 100 & 86.67\\
    \hline
    AlexNet \cite{ref35} & 100 & 100 & 100\\
    \hline
    DenseNet \cite{ref33} & 100 & 100 & 100\\
    \hline
    MobileNet \cite{ref36} & 82.76 & 85.71 & 100\\
    \hline
    \end{tabular}
\end{table}

As can be seen from the experimental results in Table \ref{Table 3}, in the digital environment :(1) the adversarial samples generated by AdvLS have very strong adversarial attack migration, which means that AdvLS have excellent performance to perform black-box adversarial attack. (2) When AlexNet \cite{ref35} is attacked by adversarial samples, the classifier was almost completely paralyzed, MobileNet \cite{ref36} and VGG19 \cite{ref34} are also almost paralyzed, while Inception\_V3 \cite{ref38} shows excellent classification performance. On the other hand, according to the experimental results in Table \ref{Table 4}, in the physical environment :(1) The adversarial samples generated by AdvLS also have strong adversarial attack migration. (2) When AlexNet \cite{ref35} and DenseNet \cite{ref33} are attacked by physical adversarial samples generated by AdvLS, they are completely paralyzed, and ResNet101 \cite{ref31} is also almost completely paralyzed. In general, AlexNet \cite{ref35} has almost no robustness to adversarial samples generated by AdvLS, while Inception\_V3 \cite{ref38} has better robustness.

The experimental results in Table \ref{Table 1} and Table \ref{Table 2} show that AdvLS has robust adversarial effect in both digital and physical environments, which means that AdvLS has a non-negligible adversarial effect in white-box conditions. The experimental results in Table \ref{Table 3} and Table \ref{Table 4} show that AdvLS has strong adversarial attack migration in both digital and physical environments, which means that AdvLS is effective to conduct black-box attacks. The experimental results of this work show that AdvLS has robust adversarial attack capability and attack migration under different environments, AdvLS conduct robust adversarial attack under white-box condition and black-box condition. In a nutshell, AdvLS pose a significant security threat to the advanced vision-based systems, so we call for AdvLS to receive widespread attention.

\section{Conclusion}
\label{sec6}

In this paper, we propose a light-based physical attack, AdvLS, which performs adversarial attack by optimizing the physical parameters of laser spots through GA. The advantages of AdvLS include: (1) AdvLS has robust adversarial attack performance in different environments, and shows excellent adversarial attack performance in both white-box and black-box settings; (2) In the physical environment, AdvLS uses laser spots as adversarial perturbations, which allows AdvLS achieve excellent covertness; (3) AdvLS is the only light-based physical attack capable of executing attacks in the daytime. In addition, the cost of deploying AdvLS is cheap, it’s easy for an attacker to implement. The attacker performs a quick physical attack by remotely controlling the laser device. Our work shows that AdvLS poses a non-negligible security threat to many vision-based systems. In the future, light-based physical adversarial attack technology will also become a research hotspot.

In the physical world, the quantification of physical adversarial perturbations is not achievable, which is also a defect of physical adversarial attack technology so far. In the future, we will continue to study light-based physical attack (e.g. spotlight attack, shadow attack). The security of vision-based systems and applications can be further improved only when more robust and covert physical attacks are explored.


\bibliography{acml22}






\end{document}